\documentclass[preprint,12pt]{elsarticle}

\usepackage{amssymb}
\usepackage{amsmath}

\usepackage{lmodern}
\usepackage[T1]{fontenc}
\usepackage[utf8]{inputenc}
\usepackage{graphicx}
\usepackage{algorithmic}
\usepackage{algorithm}
\usepackage{subcaption}
\usepackage{booktabs}
\usepackage{multirow}
\usepackage{listings}
\usepackage[usenames,dvipsnames,svgnames,table]{xcolor}
\newtheorem{definition}{Definition}
\lstdefinestyle{myCustomMatlabStyle}{
  language=XML,
  numbers=none,
  tabsize=1,
  showspaces=false,
  showstringspaces=false
  morestring=[b]",
  morestring=[s]{>}{<},
  morecomment=[s]{<?}{?>},
  stringstyle=\color{black},
  identifierstyle=\color{darkblue},
  keywordstyle=\color{cyan},
  literate={\ \ }{{\ }}1,
  breaklines=true,
  morekeywords={owl:equivalentClass,owl:someValuesFrom}% list your attributes here
}
\journal{ }

\begin{document}

\begin{frontmatter}

\title{CMOMgen: Complex Multi-Ontology Alignment via Pattern-Guided In-Context Learning}

\author[fcul]{Marta Contreiras Silva\corref{cor1}}
\cortext[cor1]{Corresponding Author: mcdsilva@fc.ul.pt}
\author[ist]{Daniel Faria}
\author[fcul]{Catia Pesquita}

%% Author affiliation
\affiliation[fcul]{organization={LASIGE, Faculdade de Ciências, Universidade de Lisboa, Edifício C6 Piso 3, Campo Grande, 1749-016},
            state={Lisboa},
            country={Portugal}}
\affiliation[ist]{organization={INESC-ID, Instituto Superior Técnico, Universidade de Lisboa, Rua Alves Redol 9, 1000-029},
            state={Lisboa},
            country={Portugal}}

%% Abstract
\begin{abstract} % max 250 words
Constructing comprehensive knowledge graphs requires the use of multiple ontologies in order to fully contextualize data into a domain. Ontology matching finds equivalences between concepts interconnecting ontologies and creating a cohesive semantic layer. While the simple pairwise state of the art is well established, simple equivalence mappings cannot provide full semantic integration of related but disjoint ontologies. Complex multi-ontology matching (CMOM) aligns one source entity to composite logical expressions of multiple target entities, establishing more nuanced equivalences and provenance along the ontological hierarchy.

We present CMOMgen, the first end-to-end CMOM strategy that generates complete and semantically sound mappings, without establishing any restrictions on the number of target ontologies or entities. Retrieval-Augmented Generation selects relevant classes to compose the mapping and filters matching reference mappings to serve as examples, enhancing In-Context Learning.
The strategy was evaluated in three biomedical tasks with partial reference alignments. CMOMgen outperforms baselines in class selection, demonstrating the impact of having a dedicated strategy. Our strategy also achieves a minimum of 63\% in F1-score, outperforming all baselines and ablated versions in two out of three tasks and placing second in the third. Furthermore, a manual evaluation of non-reference mappings showed that 46\% of the mappings achieve the maximum score, further substantiating its ability to construct semantically sound mappings.
\end{abstract}

%% Keywords

\end{frontmatter}

\section{Introduction}
Knowledge Graphs (KGs) are an exemplary knowledge base for data-centric solutions geared towards human understanding. Incorporating ontologies into KGs places data points within domain context and imbues them with meaning, leading to an improvement on information retrieval and knowledge discovery\cite{paulheim2017knowledge}. However, since ontologies are often domain-specific, constructing a KG that provides full coverage for all necessary domains likely requires interconnecting multiple ontologies with semantically complex links to capture nuanced relationships across domains.

Integrating multiple ontologies into a single cohesive semantic layer can be achieved through Ontology Matching (OM), where algorithms find correspondences (or mappings) between entities of two or more ontologies\cite{euzenat2013}.
The state of the art in OM is well established for simple equivalence mappings between pairs of related ontologies\cite{oaei2024,aml,logmap}. However, these mappings are unable to provide full semantic integration of multiple ontologies when these have fundamentally differing points of view or when they cover complementary subsections of the domain\cite{ritze2009}. This challenge can be addressed by combining multiple entities into logical expressions in the form of complex mappings.

In complex mappings, an ontology entity is mapped to an expression that is composed of one or more entities from one or more ontologies and includes at least one logical construct (e.g., an existential restriction, an intersection, etc). Due to this, complex mappings can better reconcile ontologies and connect them, since the ability to establish complex relationships between disjoint but related concepts allows knowledge to be structured as a sum of parts, enabling the connection to complex concepts established in other ontologies. For example, the concept of "\textit{increased heart rate}" can be equivalent to the expression "\textit{increased} AND \textit{heart rate}", effectively connecting all three entities in the KG. While a composite expression can use entities from a single target ontology, often the logical expression utilises more than one target ontology, leading to the paradigm of Complex Multi-Ontology Matching (CMOM).

In CMOM, the equivalences are established in a 1:n scenario, connecting an entity from the source ontology to a construct that combines multiple entities that originate from one or more target ontologies. While tasks that match multiple ontologies are often addressed by breaking the task into sets of pairwise matching tasks, in CMOM, the set of target ontologies must be considered simultaneously, precluding the creation of multiple smaller tasks to handle the multi-ontology scenario.

CMOM presents a unique set of challenges in both scalability and in achieving the final logical construct through automated methods. Devising an approach for CMOM that establishes no limits or set patterns implies a search for any number of target entities from any number of target ontologies, with boundless combinations within the expressivity limits of logical constructs. Moreover, the combination of entities into a final expression must not only be accurate in terms of the concepts it uses but also in their order, combination, and nesting. To the best of our knowledge, existing complex mappings in ontologies were achieved through manual efforts by experts, meaning new complex mappings are reliant on the availability of domain experts and signify a large investment in time and effort.

\begin{figure}[!h] {
\centering\tiny\sffamily
\begin{subfigure}{\linewidth}
    \includegraphics[width=\linewidth]{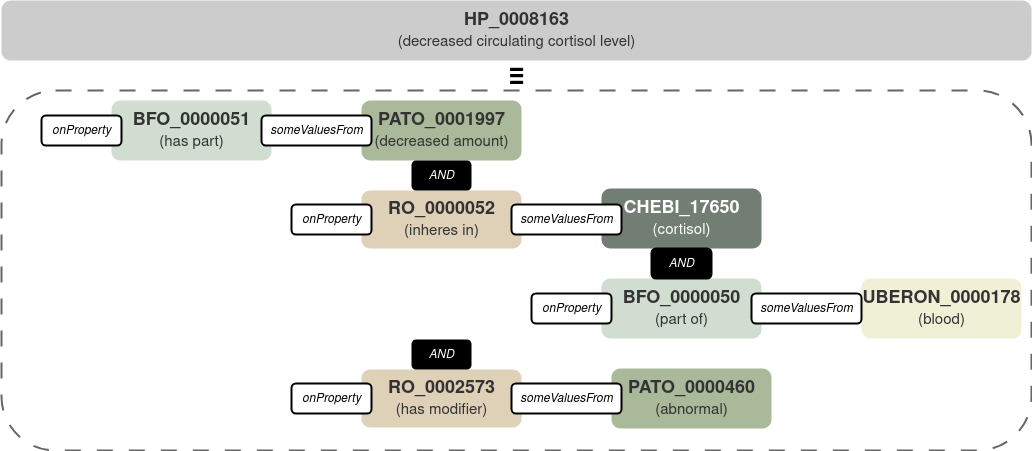}
\end{subfigure}
\lstset{language=XML}
\begin{lstlisting}
<owl:Class rdf:about="http://purl.obolibrary.org/obo/HP_0008163">	<!-- decreased circulating cortisol level -->
 <owl:equivalentClass>
  <owl:Restriction>
   <owl:onProperty rdf:resource="http://purl.obolibrary.org/obo/BFO_0000051"/>	<!-- has part -->
    <owl:someValuesFrom>
     <owl:Class>
      <owl:intersectionOf rdf:parseType="Collection">
       <rdf:Description rdf:about="http://purl.obolibrary.org/obo/PATO_0001997"/>	<!-- decreased amount -->
       <owl:Restriction>
        <owl:onProperty rdf:resource="http://purl.obolibrary.org/obo/RO_0000052"/>	<!-- inheres in -->
         <owl:someValuesFrom>
          <owl:Class>
           <owl:intersectionOf rdf:parseType="Collection">
            <rdf:Description rdf:about="http://purl.obolibrary.org/obo/CHEBI_17650"/>	<!-- cortisol -->
             <owl:Restriction>
              <owl:onProperty rdf:resource="http://purl.obolibrary.org/obo/BFO_0000050"/>	<!-- part of -->
              <owl:someValuesFrom rdf:resource="http://purl.obolibrary.org/obo/UBERON_0000178"/>	<!-- blood -->
            </owl:Restriction>
          </owl:intersectionOf>
         </owl:Class>
        </owl:someValuesFrom>
       </owl:Restriction>
       <owl:Restriction>
         <owl:onProperty rdf:resource="http://purl.obolibrary.org/obo/RO_0002573"/>	<!-- has modifier -->
         <owl:someValuesFrom rdf:resource="http://purl.obolibrary.org/obo/PATO_0000460"/>	<!-- abnormal -->
       </owl:Restriction>
      </owl:intersectionOf>
     </owl:Class>
    </owl:someValuesFrom>
  </owl:Restriction>
 </owl:equivalentClass>
</owl:Class>
\end{lstlisting}
}
 \caption{The logical definition of HP\_0008163 (decreased circulating cortisol level) in both a simplified graphical version and the original OWL construct which we replicate with CMOMgen. The OWL version includes labels in the form of comments in the same line as the respective entity.}
    \label{fig:LD}
\end{figure}

A notable example of complex mappings appears in the biomedical domain in the form of logical definitions. These represent a manual effort by ontologists to semantically define complex concepts in one ontology (e.g., phenotype) using simpler concepts from other ontologies (e.g., an anatomical structure and a quality)\cite{mungall2011}. 
An example of a logical definition is: \textit{decreased circulating cortisol level} $\equiv$ (\textit{has part} SOME (\textit{decreased amount} AND (\textit{inheres in} SOME (\textit{cortisol} AND (\textit{part of} SOME \textit{blood}))) AND (\textit{has modifier} SOME \textit{abnormal}))) shown in Figure~\ref{fig:LD}. These real examples show not only the applicability of these complex equivalences but also underline the potential of a CMOM strategy in establishing these real-life composite expressions through automated efforts in order to decrease the time and effort expended by domain experts.

The biomedical domain presents as an interesting use case for the CMOM paradigm, beyond the existence of potential reference mappings, as complex mappings in this domain can provide impact in real-life solutions, such as the diagnosis of phenotypes. Moreover, this domain demonstrates a profusion of ontologies in its various applications, resources, and data, with a very diversified use of various domain ontologies, underlining the importance of using multiple ones in self-contained solutions.

CMOM hasn't received much attention over the years, likely due to its complexity, with the state of the art being composed of very few works when compared to pairwise ontology matching. Moreover, existing strategies are often limited to specific scenarios and cardinalities, with their performances falling short of the results achieved in 1:1 alignment tasks\cite{oaei2024}.

Oliveira and Pesquita\cite{oliveira2018improving} addressed the simplest form of the problem with only two target ontologies and a fixed pattern.
Toro \textit{et al.}\cite{toro2024dynamic} developed an approach based on language models for ontology generation that aims to produce logical definitions but is restricted to simple \textit{genus-differentia} forms, which are restricted to a 1:2 cardinality and limited in complexity.
Silva \textit{et al.}\cite{silva2024ecai} focused on finding candidate target entities through a combined strategy of lexical and language model-based approaches, without producing a complete complex construct.  

We present CMOMgen, the first end-to-end approach to generate semantically sound and complete complex multi-ontology mappings.
Retrieval-Augmented Generation (RAG) provides sets of selected target classes, found through a combined lexical and language model-based strategy, and examples from reference logical definitions. This, in turn, guides In-Context Learning (ICL) to generate complex OWL expressions using a hybrid neurosymbolic AI approach.
Additionally, we present a novel evaluation for CMOM based on semantic graph distance that accounts for all its constituents.

The main contributions of this work are CMOMgen, an end-to-end approach for CMOM, a set of evaluation metrics tailored to the CMOM scenario (semantic precision, recall, and graph edit distance), as well as partial reference alignments for evaluating CMOM.

This paper follows a structured layout. Section~\ref{sec:prob} outlines the definitions used throughout this work that pertain to the complex multi-ontology matching problem. Section~\ref{sec:related} expands on the related work and the previous efforts in solving CMOM and/or using language models in ontology matching. Section~\ref{sec:methods} explains all the methods used within our strategy divided into its sequential sections. Section~\ref{sec:design} describes the experimental design used to run our approach, including: the ontologies and references used, the types of evaluation conducted (automatic and manual), the graph edit distance metric used for automatic evaluation, the baselines and ablation studies used to compare to our strategy, and the implementation. Section~\ref{sec:results} presents and discusses the results obtained from using CMOMgen in the experiments outlined in the previous section. Finally, Section~\ref{sec:conclusions} concludes this work with some final remarks, future work, and potential limitations.

\section{Problem Definition}\label{sec:prob}
Ontology matching establishes equivalences (or mappings) between a source ontology and a target ontology\cite{euzenat2013}, as per Definition~\ref{def:OM}. Mappings can be established between any type of entity, e.g., classes, properties, or individuals.

\begin{definition}[Ontology Matching]
\label{def:OM}
    An alignment between source ontology $S$ and a target ontology $T$ is a set of mappings represented as tuples in the form $<e_1, e_2, r, c>$, where $e_1 \in S$ is a source entity mapped to target entity $e_2 \in T$, $r$ is the established semantic relation between them (e.g. $\equiv$, $\geq$, $\leq$, $\bot$), and $c$ is an optional confidence score.
\end{definition}

In the complex ontology matching paradigm, mappings can include logical expressions that combine multiple entities of the target ontology in place of $e_1$ and/or $e_2$\cite{ritze2009}. In this work we focus on the 1:n variant of complex ontology matching, as per Definition~\ref{def:COM}. These mappings are able to capture more precise semantic relations, by establishing logical expressions that link multiple entities and make use of logical constructs such as intersections and union of classes, as well as restrictions on domain, range, cardinality or occurrence of properties.

\begin{definition}[Complex Ontology Matching]
\label{def:COM}
Let $S$ and $T$ be a source and a target ontology respectively, a complex 1: n mapping is represented as a tuple in the form $<e_1, \{t_1, ..., t_i\}, r, c>$, where $e_1 \in S$ is a source entity mapped to the logical expression $\{t_1, ..., t_i\} \in T$ in which the target entities are combined within the limits of OWL expressivity, $r$ is the established semantic relation between them (e.g. $\equiv$, $\geq$, $\leq$, $\bot$), and $c$ is an optional confidence score.
\end{definition}

As an extension of this paradigm, in CMOM the complex expressions are constructed using entities from multiple target ontologies, as per Definition~\ref{def:CMOM}.

\begin{definition}[Complex Multi-Ontology Matching]
\label{def:CMOM}
Let $S$ be a source ontology and $\mathcal{T}$ the set of target ontologies $\{T_1, ..., T_n\}$ with n$\geq$2. A complex 1:n multi-ontology mapping is represented as a tuple in the form $<e_1, \{t_1, ..., t_i\}, r, c>$, where $e_1 \in S$ is a source entity mapped to the logical expression $\{t_1, ..., t_i\} \in \mathcal{T}$ in which the target entities are combined within the limits of OWL expressivity, $r$ is the established semantic relation between them (e.g. $\equiv$, $\geq$, $\leq$, $\bot$), and $c$ is an optional confidence score.
\end{definition}

Solving the 1:n CMOM problem implies boundless combinations of entities from an arbitrary number of target ontologies into a logical expression that conforms to domain knowledge, which is entirely non-trivial, and has yet to be solved by existing solutions\cite{oliveira2018improving, zhou2018complex,silva2024ecai}.

Our approach, CMOMgen, provides an end-to-end strategy that handles any arity of target ontologies and produces semantically sound and complete logical constructs in the form of equivalence mappings.

\section{Related Work}\label{sec:related}

Simple pairwise matching problems are still the focus of most OM approaches, with successful tools relying on rule-based algorithms primarily based on the lexical component of ontologies, with additional support from their structural component and exploitation of background knowledge sources\cite{aml,logmap}. 

With this focus on the lexical component in ontology matching strategies, a recent ally in the development of new strategies has been language models (LMs), due to their capability of incorporating textual context in their representation and potential for fine-tuning. In the 2024 edition of the Ontology Alignment Evaluation Initiative, more specifically in its Bio-ML track, five systems used LMs to tackle the tasks proposed by this track: BERTMap\cite{he2022bertmap}, BioGITOM\cite{biogitom}, BioSTransMatch\cite{biostransmatch}, HybridOM\cite{hybridom}, Matcha\cite{matcha}. While all these systems overlap in using cosine similarity as the comparison metric between generated embeddings, the strategies and models used all vary, demonstrating how diverse the application of language models in ontology matching can be.

Some of the strategies used to find simple mappings have been adapted to the complex pairwise paradigm, which can be divided into two major categories\cite{thieblin2020survey}: instance-based strategies\cite{zhou2019towards}, which rely on the analysis of patterns between classes and properties associated with instances shared between the ontologies; and lexical-based strategies\cite{ritze2009}, which rely on finding partial lexical overlaps between entities. While the former are more robust to terminology differences between ontologies, they are limited to problems where shared instances exist, which precludes them from being a solution in scenarios where the ontologies are often from related but disjoint domains and as such share no instances.

Two recent efforts in complex OM that forego the dependency on instances and similarly uses Language Models in their approach, are the works by Amini \textit{et al.}\cite{amini2024towards} and Sousa \textit{et al.}\cite{sousa2024towards}.

Amini \textit{et al.}\cite{amini2024towards} utilise a chain-of-thought prompting strategy to find the entities involved in a potential complex mapping. Modules are used to segment the ontologies, which are provided as part of a prompt to the model to inquire about the relevant entities to the reference mapping being constructed. This strategy is tested in a smaller subset of a pairwise reference task, and relies in the existence or generation of ontology modules. Moreover, the prompting strategy appears to require some level of manual tailoring in the information provided, and could potentially be more resistent to automation for large tasks. 

The approach by Sousa \textit{et al.}\cite{sousa2024towards} automatically generated SPARQL queries to modularize source and target ontologies, in order to reduce the search space. Afterwards, a few-shot learning technique is applied with a prompt that uses the previously obtained subsets of the ontologies and returns a complex alignment in EDOAL format. However, this work focuses on the pairwise paradigm and relies on an algorithm that defines relevance through established links in between entities, which precludes its use in scenarios where there are multiple distinct target ontologies.

To the best of our knowledge, there are only three efforts that partially tackle CMOM. 
Oliveira \textit{et al.}\cite{oliveira2018improving} focuses on ternary mappings, linking one source entity to two target entities, according to specific patterns, such as addition and variation. This work adapts AgreementMakerLight's\cite{aml} lexical matchers to find incremental partial matches, until the target entities achieve full coverage of the source label. 

Toro \textit{et al.}\cite{toro2024dynamic} present DRAGON-AI, which performs language model-based ontology term completion aided by Retrieval-Augmented Generation (RAG). This approach is capable of generating relationships for source entities that combine a predicate with a target entity, using an existing portion of the source entity such as a label or definition.

Both of these strategies are restricted to the simplest CMOM scenario of a pattern of two target ontologies where one is used as a modifier. However, as established by the existing biomedical logical definitions, complex biomedical mappings often do not follow this set pattern, necessitating more ontologies combined through more complex relationships. 

Silva \textit{et al.} propose a more general approach, CMOM-RS\cite{silva2024ecai}, that takes any number of target ontologies to find a set of classes that can potentially be used to construct a complex mapping for a source entity. This approach combines two recursive strategies: a lexical one, where the target labels provide full coverage of the source label without overlap, and a language model (LM)-based one that uses geometric operations on label embeddings. CMOM-RS is, as far as we know, the only approach to CMOM with no restriction on the arity of the mapping. However, it finds only the candidate target classes to use in the complex mapping, without producing a complete logical construct.

CMOMgen integrates CMOM-RS into a full end-to-end strategy capable of generating the complete logical construct of a complex mapping using any number of target ontologies.

\section{Methods}\label{sec:methods}
\subsection{Overview}
CMOMgen finds complex mappings in OWL format represented as \\ $<e_s, \{e_{t_1},..., e_{t_m}\}, \equiv>$, where $S$ is a source ontology with $e_s \in S$ being the source entities, $e_t \in \{T_1,...,T_n\}$ are the target entities present in a set of $n$ target ontologies with $n\geq$1, which are combined using OWL constructs into a single logical expression, and the third element of the tuple corresponds to the relation captured by the mapping, in this case, equivalence.

\begin{figure}[h!]
    \centering
    \includegraphics[width=1\linewidth]{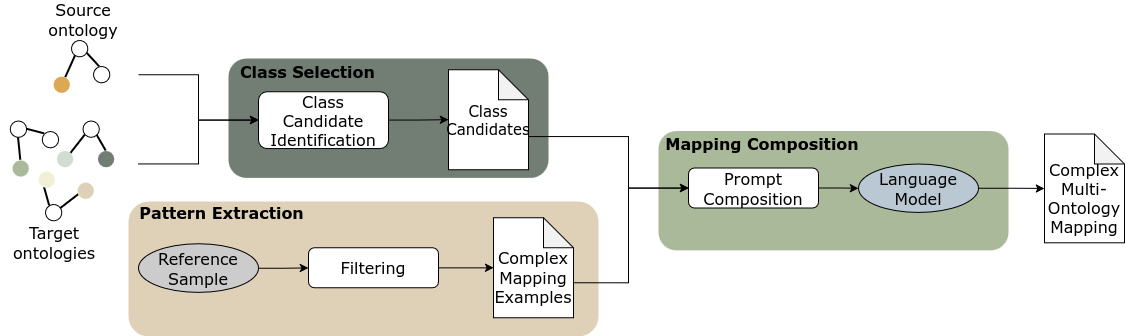}
    \caption{
    Overview of the methodology for complex multi-ontology matching.}
    \label{fig:full}
\end{figure}

A general overview of our method is summarized in Figure~\ref{fig:full}.
It takes as input a source ontology $mathcal{S}$, a set of target ontologies $\mathcal{T} \in T$, and a sample of reference complex mappings $\mathcal{R}$, i.e., equivalence class statements defined for $\mathcal{S}$. It functions as a local alignment strategy, where the goal is not to produce a final alignment but rather individual mappings for given source classes. For each source class $s_i \in \mathcal{S}$, classes selection finds the target entities $\{t_1, ..., t_n\} \in \mathcal{T}$ to combine in the expression, pattern extraction selects the reference complex mappings $\{r_1, ..., r_n\} \in \mathcal{R}$ that serve as an example set, and mapping composition prompts a language model with the previous pieces to obtain a final complex mapping $m$ that is saved individually for each $s_i$.

The classes selection relies on two complementary recursive approaches, a lexical strategy that finds non-overlapping target labels that achieve full coverage of the source label, and a language model-based that uses recursive subtraction of embeddings to find the most similar target combination to the source embedding, to be used as the selected classes in the final construct.

Our mapping composition method incorporates a Retrieval-Augmented Generation (RAG) framework where selected classes and relevant complex multi-ontology mappings are retrieved to serve as examples for in-context learning.  RAG is a language model framework that can be used to improve generated results by incorporating external information into the querying phase\cite{rag}. With in-context learning, examples of the expected results are provided to the language model in order to direct it towards improved results, directly through the query without updating the model\cite{icl}. 

The overall approach is composed of the following steps: (1) extraction and pre-processing of the ontology vocabularies; (2) classes selection, through two complementary approaches: one lexical-based and one language model-based; (3) aggregation and filtering of the sets of selected classes; (4) mapping pattern extraction from existing complex multi-ontology mappings that fit the selected classes set; (5) mapping composition using in-context learning that leverages the selected classes and the extracted patterns.

\subsection{Pre-processing}
CMOMgen is vocabulary-based as it relies only on the names and synonyms established in the ontologies. As such, it is of the utmost importance to maximize the names available to our strategy by pre-processing ontologies' lexicons into unified data structures. The set of labels from the source ontology are extracted into a source vocabulary and the set of labels from all target ontologies are extracted into a unified target vocabulary. Each label $l_i$ is associated with its corresponding class and with a confidence score $weight(l_i)$ that reflects the semantics of its label property, as the origin (whether it's a local name, synonym, formula or other) reflects the proximity of the label to its entity. The confidence score $weight(l_i)$ of each label $l_i$ follows:

\begin{equation}
\label{eq:lex}
weight(l_i) = \begin{cases} 
1 & \text{if local name} \\
0.95 & \text{if label} \\
0.9 & \text{if exact synonym or internal synonym} \\
0.85 & \text{if other synonym or external synonym} \\
0.8 & \text{if formula} \\
\end{cases}
\end{equation}

This score is adjusted according to the number of names of an entity that share the same origin, in order to penalize instances where there is a large number of labels from the same origin, potentially indicating a less rigorous selection of synonyms. The corrected score is calculated for each label $l_i$ following:

\begin{equation}
\label{eq:adjust}
   \text{corrected weight}(l_i) = \text{max}(weight(l_i)-\frac{|L|}{100}, 0)
\end{equation}

where $|L|$ corresponds to the total number of labels that have the same semantic property as $l_i$ and is divided by the empirically defined value of 100.

\begin{figure}[h!]
    \centering
    \includegraphics[width=\linewidth]{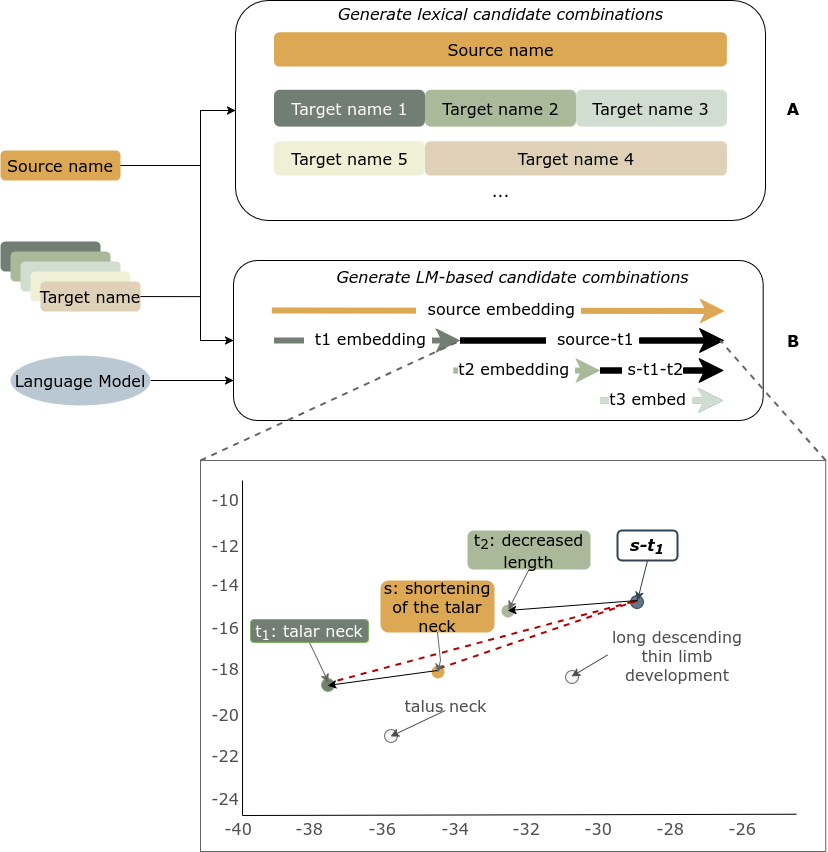}
    \caption{
    Classes selection through two complementary strategies. (A) Lexical classes selection recursively finds target names that provide full coverage of the source name without overlap. (B) Language model-based classes selection uses recursive subtraction of the target name embeddings against the source name embeddings.}
    \label{fig:candidates}
\end{figure}

\subsection{Classes Selection}
The first step of our method identifies, for a given source class $e_s$, the target classes $E_c={e_{t_1},..., e_{t_m}}$ that will be combined into a complex mapping downstream. The strategy adopted was introduced in~\cite{silva2024ecai} and is composed of two complementary recursive strategies which are summarized in Figure~\ref{fig:candidates}. This approach affords desirable properties by imposing no restriction on the arity of the mappings and exploring the complementarity between a recall-oriented approach based on language embeddings and a precision-oriented approach based on lexical similarity.

\subsubsection{Lexical Classes Selection}
For each source name, all target names that share at least one word are selected. A recursive function uses this filtered target name set to compose all possible combinations of names that provide full coverage of the source name without any overlap between themselves. As seen in Figure~\ref{fig:candidates}A, the source name is entirely covered without overlap by the combinations $\{t_1\text{name}, t_2\text{name}, t_3\text{name}\}$ and $\{t_5\text{name}, t_4\text{name}\}$, originating two sets of selected classes. Several valid combinations can be found for one source entity and the confidence score of each selected set is calculated as the product of the confidence of each target name $n_{t_j}$ within it (see Equation~\ref{eq:lex-score}).

\begin{equation}
\label{eq:lex-score}
score=\prod_{j=1}^{n} conf(n_{t_j})
\end{equation}

\subsubsection{Language Model-based Classes Selection}
Unlike the lexical strategy, the language model-based strategy generates a single target combination for each source name by exploring geometric operations of vectorial representations of the names. This recursive approach finds the most similar target name embedding to a source name embedding, updates the source embedding by subtracting the target embedding, and recursively finds the next most similar target embedding to the updated embedding. As seen in the detail of Figure~\ref{fig:candidates}B, the most similar target embedding to the source embedding $s$ for "shortening of the talar neck" is $t_1$ "talar neck". The subtraction between these two embeddings is marked as $s-t_1$, and the most similar target embedding to this subtraction embedding is $t_2$ "decreased length". In the end, the selected classes set for $s$ is $\{t_1, t_2\}$.

The recursive function stops once the cosine similarity between the two embeddings is lower than the input parameter $\alpha$. The final class set is composed of the target names found through this approach for each source name. 
More formally, let $s$ be the source name embedding, then the best mapping for $s$ can be found following Equation~\ref{eq:map}.

\begin{equation}
\label{eq:map}
map(s) = \begin{cases} 
M = M \cup {t_{max}} & \text{if } cos(s,t_{max}) < \alpha \\
map(s-{t_{max}}) & \text{if } cos(s,t_{max}) \geq \alpha\\
\end{cases}
\end{equation}

where $M$ is the set of target names selected to compose the best mapping, $s-{t_{max}}$ corresponds to the subtraction of the vectors $s$ and $t_{max}$, where $t_{max}$ is the embedding for the most similar target name, computed by maximizing the cosine similarity between the target names embeddings $t_i$ and the source name embedding (see Equation~\ref{eq:t-max}).

\begin{equation}
\label{eq:t-max}
t_{max} = \underset{\mathbf{t}_i}{\text{argmax}} \, cos(s,t_{i})
\end{equation}

The confidence score for each set of selected classes is calculated as the cosine similarity between the sum of the target name embeddings $t_j$ within it and the source name embedding $s$ (see Equation~\ref{eq:conf_score}).

\begin{equation}
\label{eq:conf_score}
\text{score} = \text{cos}(\mathbf{s}, \sum_{j=1}^{n} \mathbf{t}_j)
\end{equation}

\subsection{Aggregation and Filtering}
Both sets of selected classes, lexical and LM-based, are aggregated into a single alignment of approximately 1:1 through a greedy heuristic that selects in descending order of confidence score. Any tied sets are all returned, following the permissive selection paradigm proposed in\cite{aml}. The selected sets are used for the pattern extraction by providing the general pattern for the selection of examples.

\subsection{Pattern Extraction}
The general pattern of the selected classes' sets is used to filter examples for in-context learning.
This pattern focuses on the target ontologies used (using namespaces as an approximation) and their cardinality.

For each set of selected target classes $M$, the namespaces of their target ontologies $\mathcal{T}$ are extracted maintaining their respective cardinality, meaning if a set has two entities from $T_1$ and one from $T_2$, the namespace set will reflect that as $\{T_1\text{namespace}, T_1\text{namespace}, T_2\text{namespace}\}$. The namespace set is then used to filter the complex equivalence class statements ($r \in \mathcal{R}$) to remove statements that do not match the namespaces in their respective cardinality. They can, however, include more entities from the same or other target ontologies. More formally, 

\begin{equation}
\mathcal{R}_\text{filtered} = \{ r \in \mathcal{R} \mid \text{namespace}(r) \subseteq \text{namespace}(E_c)
\end{equation}

\subsection{Mapping Composition}
To obtain a complex mapping for each source entity, a language model is prompted to construct this mapping using the knowledge previously retrieved. Prompt composition takes as input the IRI and main label of the source entity, the IRIs and main labels of the entities in the selected classes set, and the mapping examples filtered from the logical definitions in their original OWL format. The LM should return an OWL expression that is equivalent to the source class provided and follows the structure and semantics of the examples using the classes given. This approach follows a hybrid neurosymbolic AI approach\cite{pan2023large} whereby semantic knowledge extracted from the ontologies informs the neural language model through in-context learning.

\subsubsection{Prompt Engineering}
The prompts were constructed with three objectives in mind: ensure the construct was in parseable OWL format, that it used an \textit{equivalentClass} statement and that it used entities from the Relation Ontology (RO) and the Basic Formal Ontology (BFO) to mimic existing reference mappings. The querying was conducted through the API, with the prompt being split into two roles, one of \textit{system} that provides guidelines on what is the goal of the task and the general constraints that apply to it, and \textit{user}, which includes the specific information about the task, meaning the source entity, selected classes and mapping examples. Prompts were adjusted empirically on a subset of 15 complex ontology mappings from the HP ontology, corresponding to the most common patterns of logical construct regardless of target namespace. The \textit{system} message was the same in all prompts: 

\textit{``You will receive a request to construct a complex mapping in OWL format. You have to answer with an ontology in OWL format that can be read by rdflib, no explanations. Make sure to use an equivalentClass statement." }\footnote{The mention of python package rdflib (https://github.com/RDFLib/rdflib) was an attempt to force parseable OWL expressions.}

The \textit{user} message varied according to the type of query and was constructed by composing the different excerpts  in Table~\ref{tab:prompts}: initial request, selected classes, examples, property ontologies requirement.

\begin{table}[h!]\centering
\caption{Prompt excerpts used to compose each type of query.}
\footnotesize
\begin{tabular}{m{3cm}m{10cm}}
\toprule
Type & Template \\
\midrule
Initial request & ``Create a complex mapping in OWL format for the class [\underline{class IRI}] ([\underline{main class label}])." \newline OR \newline ``Create a complex mapping in OWL format for the class [\underline{class IRI}] ([\underline{main class label}]) according to the examples provided."  \\
\midrule
Selected classes & ``You should use the following classes: [\underline{list of selected classes with IRI and main label}] and any others you find necessary to match the appropriate pattern in the examples."\\
\midrule
Examples & ``This is a list of possible equivalentClass examples for complex mappings: [\underline{list of examples}]."\\
\midrule
Property ontologies requirement & ``You are only allowed to use properties from the Relation Ontology (RO) and the Basic Formal Ontology (BFO)." \\
\bottomrule
\end{tabular}
\label{tab:prompts}
\end{table}

\section{Experimental Design}\label{sec:design}
\subsection{Ontologies and Reference Alignments}\label{sec:ontos}
We evaluated CMOMgen in three biomedical tasks which focus on different source ontologies: the Human Phenotype Ontology (HP)\cite{hp}, the Mammalian Phenotype Ontology (MP)\cite{mp}, and the Worm Phenotype Ontology (WBP)\cite{wbp}. 
These ontologies describe phenotypic characteristics, which can be broken down into an expression where an anatomical part or biological component is altered by a particular quality, e.g. "aplasia of the bladder" (HP:0010477) being equivalent to an expression that combines "urinary bladder" (UBERON:0001255) with "aplastic" (PATO:0001483). Encoding a concept as a sum of simpler concepts maintains the ability to traceback to these simpler concepts whenever necessary, such as being able to link diverse diseases by the anatomical part that they affect (e.g. urinary bladder in the previous example).
The construction of complex biomedical mappings holds potential for real-world impact as they can help describe, trace, and diagnose phenotypes in multiple species.

The logical definitions within these ontologies are encoded as \textit{equivalentClass} statements linking a source class to a logical expression composed of multiple target classes which can belong to any number of target ontologies. The logical definitions were filtered in order to focus on the set of target ontologies that were used more often (see Table~\ref{tab:ontos}), with this final set comprising the base for the partial reference alignment used to perform the reference-based evaluation of CMOMgen.
The reference for both HP and MP was constructed with the following target ontologies: the Cell Ontology (CL)\cite{cl}, the Chemical Entities of Biological Interest (ChEBI)\cite{chebi}, the Gene Ontology (GO)\cite{go}, the Phenotype and Trait Ontology (PATO)\cite{pato}, and the Uber Anatomy Ontology (UBERON)\cite{uberon}. The WBP reference, uses: ChEBI, GO, PATO, and the \textit{C.elegans} Gross Anatomy Ontology (WBbt)\cite{wbbt}.

\begin{table}[h!]\centering
\caption{Frequency of usage of unique entities in each target ontology in the logical definitions of the three source ontologies and number of total mappings extracted}
\footnotesize
\begin{tabular}{llrrr}
\toprule
 & &\multicolumn{3}{c}{Source ontologies} \\
Target ontologies & & HP & MP & WBP \\
 \cmidrule{1-1}
  \cmidrule{3-5}
    ChEBI\cite{chebi} & & 219 & 172 & 110 \\
    CL\cite{cl} & & 82 & 363 & 5 \\
    GO\cite{go} & & 237 & 834 & 305 \\
    PATO\cite{pato} & &308 & 369 & 103 \\
    UBERON\cite{uberon} & &1154 & 2093 & 1 \\
    WBbt\cite{wbbt} & &- & - & 136 \\
\midrule
Total mappings&&5447 & 9311 & 869 \\
\bottomrule
\end{tabular}
\label{tab:ontos}
\end{table}

\subsection{Reference-based Evaluation}\label{sec:ref-based}
Partial reference alignments were created for each of the three tasks, composed of the logical definitions of each of the three source ontologies that used entities from the selected target ontologies, as outlined in Section~\ref{sec:ontos}. These partial reference alignments were used to conduct a reference-based evaluation of CMOMgen, using the existing mappings as the reference mappings.
This evaluation is divided into two major sections: a class-based that simply takes into account the classes present in the final construct (Section~\ref{sec:class-based}); and a graph-based one, where the complete construct is taken into account with its logical expression (Section~\ref{sec:ged}). 
Additionally, to further assess the impact of our full strategy against simpler approaches, we established two baselines (one based on the classes selection and another based on a simpler prompting strategy) and constructed two ablation studies, where versions of our prompt were created to evaluate the influence of each of its parts (Section~\ref{sec:ablation}).

\subsubsection{Class-based Evaluation}\label{sec:class-based}
As classes selection is an essential part of the CMOMgen methodology, we have assessed the correctness of the classes used in the final construct. 

We conducted a detailed class evaluation as used by us in \cite{silva2024ecai}, where partial correctness is taken into account. 
With this evaluation, the relaxed precision, recall, and f-measure for each selected target class is calculated taking into account the semantic relation to the reference entity as sub or superclass, as a related class represents a partially correct match that can be easily corrected by a domain expert.

Let $e_i$ and $e_{rj}$ be a predicted entity and a reference entity, respectively, and $>$ and $<$ stand for direct sub- or super-classes, then for each target entity the relaxed precision is calculated following:

\begin{equation}
\label{eq:relax_prec}
relaxed~prec(e_i)=
\begin{cases}
1 & \text{if } e_i \leq e_{rj} \\
0.5 & \text{if } e_i > e_{rj} \\
0 & \text{otherwise } \\
\end{cases}
\end{equation}

and the relaxed recall is calculated following:

\begin{equation}
\label{eq:relax_rec}
relaxed~rec(e_i)=
\begin{cases}
1 & \text{if } e_i \geq e_{rj} \\
0.5 & \text{if } e_i < e_{rj} \\
0 & \text{otherwise } \\
\end{cases}
\end{equation}
\noindent

The precision of the complete class set $m_i$ is calculated as the ratio between the sum of each target's relaxed precision and the number of targets in the set, following:

\begin{equation}
    \label{eq:relax_map_prec}
    prec(m_i) = \frac{\sum_{i=1}^{n} \text{relaxed precision}(e_i)}{n}
\end{equation}

with $n$ being the number of target entities in the set, and $m_i$ the set of selected classes $\{e_1,.., e_n\}$.

Recall of set $m_i$ is calculated with a ratio between the sum of each target's relaxed recall and the number of targets in the selected classes set, following:.

\begin{equation}
    \label{eq:relax_map_rec}
    rec(m_i) = \frac{\sum_{i=1}^{n} \text{relaxed recall}(e_i)}{r}
\end{equation}
where $r$ stands for the number of target entities in the reference set.

The precision and recall of each task is presented as the average of all selected classes sets.

\subsubsection{Graph Edit Distance}\label{sec:ged}
To construct a complete complex mapping, the correct classes have to be selected and combined in a manner that is correct within domain knowledge and accurately translates into the source concept. As such, the comparison to a reference logical definition must also take into account not only the presence or absence of reference classes but also their order and the logical connectors that link them.
A proper evaluation of our full method requires the evaluation of the correctness of the entire \textit{equivalentClass} expression, which can include, in addition to any number of target entities, connectors such as \textit{someValuesFrom}, \textit{intersectionOf}, or \textit{unionOf} combined within the expressivity of the OWL language.

To tackle this challenge, we developed a novel evaluation approach based on graph edit distance that extends previous contributions in the semantic evaluation of both simple and complex alignments 
\cite{ehrig2005relaxed,zhou2019towards,silva2024ecai,santos2025evaluation}. Particularly, our approach accounts for the effort required for correcting the mapping and considers the semantics of the mapping components unlike recent efforts\cite{santos2025evaluation}.

\begin{figure}[h!]
\centering
\includegraphics[width=0.7\linewidth]{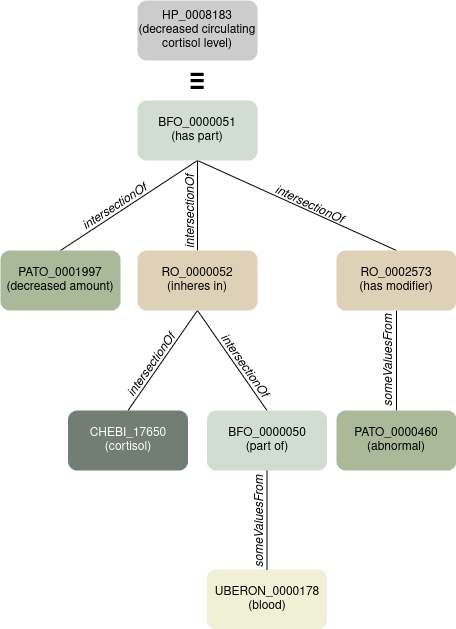}
\caption{Example of the graph transformation of the mapping in Figure~\ref{fig:LD}.}
\label{fig:map2graph}
\end{figure}

A generated mapping for a given source entity is considered fully correct if it is syntactically equivalent to the reference mapping and partially correct if it is similar to it. To assess similarity,  mappings are transformed into graphs, and their semantic graph edit distance is computed. The process of transformation maps each class or property into a node identified by its IRI, and OWL class expression constructs are captured as edges identified by their name. Figure~\ref{fig:map2graph} illustrates a graph transformation for the mapping in Figure~\ref{fig:LD}.

The semantic graph edit distance finds the minimal cost of transforming the generated mapping into the reference mapping. It considers node and edge insertions, deletions, and substitutions, which are weighted using an approach that captures the effort required to correct the mapping.

Consider that $N$ is the total number of classes in all target ontologies, that $N(c)$ is the number of direct neighbors of entity $c$ in its ontology (superclass and subclass), and that $<$ and $>$ denote direct superclass and direct subclass axioms. Then, the substitution cost of a node ($n_c$) for its corresponding reference node ($n_r$) is captured by the effort required to correct it. 
The worst case, corresponding to a cost of 1,  requires considering all entities in the target ontologies. However, we consider that in a realistic scenario, a human validator would first consider the direct neighborhood of $n_c$, significantly decreasing the effort. As such, the cost of substituting a node is formalized as Equation~\ref{eq:node_sub_class}, where $\sigma$ is a scaling factor to reflect that the correcting effort differs according to the type of entity being corrected.

\begin{equation}
\label{eq:node_sub_class}
    node_{sub}(n_c, n_r) = \\
    \begin{cases}
        0 & \text{if } n_c = n_r \\
        \sigma \; N(n_c)/N & \text{if } n_c < n_r \vee   n_c > n_r\\
        \sigma & \text{otherwise} \\
    \end{cases}
\end{equation}

The cost of inserting and deleting an edge as well as deleting a node is based on a probabilistic definition of a binary decision of yes or no, whereas inserting a node has a maximum cost as it would imply a search over all target entities.

\begin{equation}
\label{eq:minis}
\begin{split}
    node_{ins}=1 \\ 
    node_{del}=0.5  \\
    edge_{ins/del}=0.5
    \end{split}
\end{equation}

An exact graph edit distance algorithm is employed to compute the minimal cost of transforming the generated mapping into the reference mapping\cite{abu2015exact}, which is then used to arrive at a final score $s$ for the mapping as follows:

\begin{equation}
\label{eq:ged_score}
s=1-GED(m_c,m_r)/max_{GED}
\end{equation}

where $max_{GED}$ is the potentially maximum edit distance, computed as:

\begin{equation}
\label{eq:max_edit}
max_{GED} = \begin{cases} \alpha \, max(|E_{r}|,|E_{c}|) + \beta \,|V_{r}|, \\
\hfill\text{if} |V_{c}|-|V_{r}| \leq 0 \\ 
\alpha \,  max(|E_{r}|,|E_{c}|) + \beta \,|V_{r}|+\gamma \,(|V_{c}|-|V_{r}|),\\
\hfill\text{if} |V_{c}|-|V_{r}| > 0 \\ \end{cases}
\end{equation}

where $\alpha$ is the edge cost, $\beta$ is the node substitution cost and $\gamma$ the node insertion cost.

The precision and recall of the alignment are calculated according to Equations~\ref{eq:map_prec} and~\ref{eq:map_rec}, respectively,

\begin{equation}
    \label{eq:map_prec}
    precision = \frac{\sum_{i=1}^{n} \text{s}(m_i)}{|M_c|}
\end{equation}

\begin{equation}
    \label{eq:map_rec}
    recall = \frac{\sum_{i=1}^{n}s({m_i})}{|M_r|}
\end{equation}

 where $s(m_i)$ is the score of each mapping, $|M_c|$ is the total number of mappings with valid \textit{equivalentClass} statements, and $|M_r|$ is the total number of reference mappings in the test set.

 \subsubsection{Baselines and Ablation Studies}\label{sec:ablation}
 While our strategy utilizes retrieval-augmented generation to obtain selected target classes and examples to then perform in-context learning, the model can be queried with simpler approaches that could potentially provide comparable results. Therefore, to assess the true impact of our in-context learning approach we employed two baselines: the \textbf{ (1) CMOM baseline}, which is simply the set of selected classes returned by CMOM-RS\cite{silva2024ecai}, the current state-of-the-art method in CMOM, and the \textbf{(2) LM baseline}, which corresponds to using a simple prompt that asks the LM to construct a complex mapping for the source entity. Additionally, we also conducted two ablation studies: \textbf{(3) CMOMgen without examples}, where complex mapping examples are not used for in-context learning; \textbf{(4) CMOMgen without selected classes},  where the selected classes are not used for in-context learning. In more detail and referring to Table~\ref{tab:prompts}, (2) uses only the initial requirement and property ontologies requirement; (3) uses both of these and the selected classes section, while (4) uses the two and the examples section.

These versions were compared to our full strategy through precision, recall and F1-score, as outlined in Sections~\ref{sec:class-based} and~\ref{sec:ged}.

\subsection{Expert-based Evaluation}
To assess the plausibility of non-reference mappings, i.e., mappings produced for entities not present in the available logical definitions, an OWL expert with domain experience analyzed a subsection of the constructs produced by CMOMgen.
Utilizing previous classes selection results obtained in\cite{silva2024ecai}, source classes were ranked using Equation~\ref{eq:conf_score}, which calculates a confidence score by comparing the source name embedding with the sum of the target name embeddings through cosine similarity. The top 50 were selected and the two remaining steps of CMOMgen were conducted: pattern extraction, and mapping composition.

After obtaining the constructs from the language model, these were added to a table and presented to the expert for evaluation. For each mapping, the OWL expression and its respective Manchester syntax conversion were presented alongside the label of the source entity and its definition provided by the ontology. 
The expert was asked to grade each mapping on a scale from 1 to 5 for fidelity of the complex mapping to the source entity, with 1 meaning "No Fidelity" and 5 "Full Fidelity", in order to approximate the correctness of the expression. There were no limits placed on searches for external information. A field was also added for the addition of notes, in hopes of gaining insightful comments into the plausibility of the constructs.

\subsection{Experiments}
Complex mappings were generated using our strategy for a subset of the source entities present in the partial reference alignments of each source ontology. 
For both the HP and the MP tasks, mappings were created for 1000 randomly selected logical definitions, while for WBP only 700 were created due to the smaller size of the reference.

\begin{table*}[t]\centering
\caption{Results for the class-based evaluation comparing our strategy with a baseline and one ablated version of CMOMgen. Values in bold are the highest for each task by column.}
\footnotesize
\begin{tabular}{>{\centering\arraybackslash}m{0.7cm}m{2.5cm}>{\centering\arraybackslash}m{3cm}>{\centering\arraybackslash}m{3cm}>{\centering\arraybackslash}m{3cm}}
\midrule
 &  & Semantic Precision & Semantic Recall & Semantic F1-score \\
 \midrule
\multirow{3}{*}{HP}
 & LM baseline & 0.003 & 0.001 & 0.001  \\
 & CMOMgen w/o selected classes & 0.314 & 0.342 & 0.322 \\
 & CMOMgen & \textbf{0.505} & \textbf{0.432} & \textbf{0.440} \\
 \midrule
\multirow{3}{*}{MP}
 & LM baseline & 0.013 & 0.004 & 0.005  \\
 & CMOMgen w/o selected classes & 0.431 & 0.455 & 0.437 \\
 & CMOMgen & \textbf{0.676} & \textbf{0.579} & \textbf{0.611} \\
 \midrule
\multirow{3}{*}{WBP}
 & LM baseline & 0.069 & 0.030 & 0.040  \\
 & CMOMgen w/o selected classes & 0.354 & 0.332 & 0.340 \\
 & CMOMgen & \textbf{0.463} & \textbf{0.376} & \textbf{0.404} \\
 \bottomrule
\end{tabular}
\label{tab:class-based}
\end{table*}

\subsection{Implementation}
All code is available on GitHub\footnote{https://github.com/liseda-lab/CMOMgen}. The classes selection uses the CMOM-RS approach implementation available in its repository\footnote{https://github.com/liseda-lab/CMOM-RS} and reproduced with the parameters described in\cite{silva2024ecai}. Input parameter $\alpha$ in Equation~\ref{eq:map} was set to 0.2.

Regarding pattern extraction, since all existing logical definitions in HP, MP and WBP were formulated using a default pattern that includes a final restriction of the expression \textit{has\_modifier} (RO\_0002573) SOME \textit{abnormal} (PATO\_0000460), accounting for its presence in the mappings would distort evaluation results. Due to this ubiquitousness, this PATO entity is not taken into account when extracting the namespaces and is not counted for cardinality in the examples pattern match.

Mapping composition queries the gpt-4o-mini model\footnote{https://openai.com/index/gpt-4o-mini-advancing-cost-efficient-intelligence/} through its API, using a split prompt that uses the roles \textit{system} and \textit{user}. All prompts are run with a seed of value 42 and no memory is saved between runs.

The graph edit distance was computed using NetworkX\cite{networkx} and the equations outlined in Section~\ref{sec:ged}, with the scaling factor $\sigma$ from Equation~\ref{eq:node_sub_class} set as 0.7 for classes and 1 for properties.

\section{Results and Discussion}\label{sec:results}
The results for the semantic class-based evaluation are presented in Table~\ref{tab:class-based}, comparing CMOMgen to the LM baseline and one ablated version where only the reference examples are provided in the prompt. Our complete strategy outperforms both the other prompting strategies that do not provide any target classes, underlining the importance of a dedicated classes selection method. As expected, providing the examples achieves better results as they provide guidance regarding not only the structure, but also the namespaces and general patterns of the classes used in the complex mappings.

\begin{table*}[h!]\centering
\caption{Results for the reference-based evaluation of CMOMgen against two baselines and two ablated versions.}
\footnotesize
\begin{tabular}{>{\centering\arraybackslash}m{0.7cm}m{5cm}>{\centering\arraybackslash}m{1.5cm}>{\centering\arraybackslash}m{1.5cm}>{\centering\arraybackslash}m{1.5cm}}
\midrule
 &  & Precision & Recall & F1-score \\
 \midrule
\multirow{5}{*}{HP} & CMOM baseline & 0.211 & 0.211 & 0.211 \\
 & LM baseline & 0.210 & 0.203 & 0.207  \\
 & CMOMgen w/o examples & 0.275 & 0.255 & 0.265 \\
 & CMOMgen w/o selected classes & 0.617 & 0.617 & 0.617 \\
 & CMOMgen & \textbf{0.634} & \textbf{0.632} & \textbf{0.633} \\
 \midrule
\multirow{5}{*}{MP} & CMOM baseline & 0.254 & 0.254 & 0.254 \\
 & LM baseline & 0.203 & 0.193 & 0.198  \\
 & CMOMgen w/o examples & 0.287 & 0.270 & 0.278 \\
 & CMOMgen w/o selected classes & 0.645 & 0.642 & 0.643 \\
 & CMOMgen & \textbf{0.666} & \textbf{0.662} & \textbf{0.664} \\
 \midrule
\multirow{5}{*}{WBP} & CMOM baseline & 0.217 & 0.217 & 0.217 \\
 & LM baseline & 0.206 & 0.197 & 0.201  \\
 & CMOMgen w/o examples & 0.260 & 0.242 & 0.250 \\
 & CMOMgen w/o selected classes & \textbf{0.731} & \textbf{0.730} & \textbf{0.731} \\
 & CMOMgen & 0.687 & 0.687 & 0.687 \\
 \bottomrule
\end{tabular}
\label{tab:ablation}
\end{table*}

Table~\ref{tab:ablation} presents the results of the GED-based evaluation of our method compared to all the baselines and ablated versions outlined in Section~\ref{sec:ablation}. As expected, the LM baseline has the poorest performance as it consists of providing a simple prompt to the model without additional information, and it fails in both finding the correct classes and in composing the mapping. The CMOM baseline, which corresponds simply to a set of selected classes as independent nodes,  improves on these results, even without constructing an OWL expression. The two ablated versions of CMOMgen exhibit very different behaviors. While the CMOMgen variant without mapping examples achieves a small improvement over the CMOM baseline, CMOMgen without the selected classes achieves a more substantial improvement over the baselines, demonstrating  that providing the examples results in better performance as the model returns a more coherent mapping that could be more easily edited into the final mapping. The complete CMOMgen, which provides both the selected classes and the examples, outperforms all other methods in two of the three tasks, with a more than three-fold increase in performance over the baselines. 
CMOMgen fails to outperform the ablation study without the selected classes in the WBP task, which could be explained by small errors in the selection of classes being compounded in the mapping composition step. Class-based evaluation for this task shows a smaller improvement from ablation study to CMOMgen and each erroneous class will lead to at least one wrong construct, which can exponentially aggravate any errors.

\begin{table*}[t!]\centering
\caption{Example of the manual evaluation conducted by an expert of the CMOMgen mappings.}
    \footnotesize
    \begin{tabular}{rm{9cm}}
    \midrule
        \textbf{Source entity} & HP\_0400001 \\
        \textbf{Label} & chin with vertical groove, cleft chin \\
        \textbf{Definition} & Vertical crease fold situated below the vermilion border of the lower lip and above the fatty pad of the chin with the face at rest \\
        \textbf{Manchester conversion} & EquivalentTo: has part some (chin and (inheres in some cleft)) \\
        \textbf{Fidelity} & 4 \\
        \textbf{Notes} & All the right elements, but with "cleft" and "chin" in swapped places (swapping them results in the correct mapping) \\
        \bottomrule
    \end{tabular}
    \label{tab:ex_manual}
\end{table*}

An example of the manual evaluation results can be seen in Table~\ref{tab:ex_manual}. The mapping was also presented in OWL format, however this has been omitted for space. In this example, the target classes were correctly selected and the mapping is constructed in a logical pattern. However, the target entities were placed incorrectly, needing to be switched to achieve a fully correct mapping.

\begin{figure}[b!]
    \centering
    \includegraphics[width=0.7\linewidth]{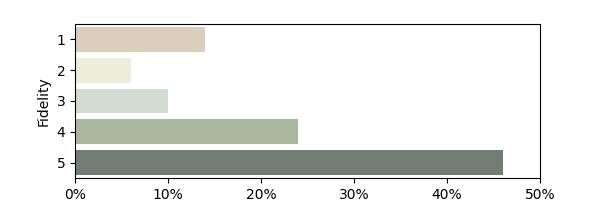}
    \caption{Distribution plot of fidelity across 50 non-reference complex mappings.}
    \label{fig:chart}
\end{figure}

The results of the expert evaluation of 50 non-reference mappings generated with our approach are shown in Figure~\ref{fig:chart}. The average score for the 50 mappings is 3.8, with 46\% of the mappings having a maximum score of 5 (signifying that they faithfully capture the semantics of the mapped concept), an additional 24\% having a positive score of 4 (signifying that they are reasonable approximations), and only 30\% of the mappings having a neutral or negative score. 

These results demonstrate that the model is able to recreate sensical constructs that accurately translate the complex concepts.
However, there are still some interesting behaviors that could be further investigated.
Sometimes the LM model mislabels IRIs in the mapping, such as when it labels PATO\_0000504 as "irregularity" when its true label is "arrhythmic", which is closer to the source term of HP\_0004308 "ventricular arrhythmia".

It is also interesting to note that the LM model will sometimes add more classes in order to complete the construct (even if it mislabels them sometimes), which could potentially fill in gaps in the class selection if further explored. For example, for HP\_0033328 "type II pneumocyte hyperplasia", the classes selected were CL\_0002063 "type II pneumocyte" and PATO\_0000644 "hyperplastic", which could potentially provide a complete mapping, but the model adds UBERON\_0002047 which it incorrectly labels as "lung" when the true label is "pontine raphe nucleus". This demonstrates a pattern where the model uses labels that are closer to the source concept than the true labels of the classes used, underlining the importance of asking for both the IRI and label.

\section{Conclusions}\label{sec:conclusions}
Complex multi-ontology matching (CMOM) provides an opportunity to improve the granularity of any ontology-based knowledge base by linking complex concepts to a construct of simpler ones, effectively connecting related but disjoint domains and maintaining traceability to all used entities throughout the linked ontologies. 

We present CMOMgen, the first end-to-end method for complex multi-ontology alignment that returns complete and semantically sound complex mappings without any restrictions on the number of target ontologies or entities and without any pre-established patterns. It enhances in-context learning with a retrieval-augmented generation framework that selects relevant classes and reference examples. CMOMgen achieves good results using language models without any fine-tuning, which translates to a more adaptable method that is not as computationally heavy.

Our reference-based evaluation demonstrated CMOMgen's ability to generate complex mappings that matched the majority of references and improved performance by three-fold over the baselines for the state-of-the-art in complex multi-ontology matching and for the language model. Further studies with two ablated versions confirmed that providing mapping examples is the most important component of the in-context learning approach. Overall, these results show the effectiveness of our method at constructing complex mappings, and underline the importance of providing external information, in particular mapping examples, to achieve improved results. 

Furthermore, an expert evaluation of the quality of non-reference mappings concluded that 46\% faithfully capture the semantics of the mapped concept, with a further 24\% being reasonable approximations. These results demonstrate that CMOMgen is able to generate semantically sound mappings and highlight its potential to extrapolate to all entities not yet covered by logical definitions.

While the results obtained are very promising, the use of language models is an inherent limitation of CMOMgen. The computational and runtime costs have yet to be addressed when working with LM solutions, which can turn users away from using these types of strategies in larger problems. Moreover, while beyond the scope of this work, the performance of the chosen model in generating mappings is crucial, more so as better models can often be unavailable, costly, or too computationally heavy for the average user.
Additionally, certain steps of the approach, such as the class selection and the example filtering, propagate errors to the next steps, which can lead to divergence, making the overall results of CMOMgen hinge on the level of performance of each step.

CMOMgen, as the first complete solution for CMOM, will benefit from the evolution in both the ontology matching and the language model domains. 
There is also an opportunity to build upon these results and seek additional refinement, for example, by further testing parameters, weights, and scores in a variety of problems. 
While CMOMgen was designed as an answer to CMOM, its use in other paradigms, such as complex pairwise alignment, must be investigated, as it can potentially stand as an adaptable strategy to multiple OM scenarios.
Moreover, its performance in other domains should also be explored since the present work focused solely on phenotype ontologies. Regarding this exploration, however, the limiting factor will be the existence of partial references that can be used to assess its results.

As a step forward in automation, it is important to note that most efforts to construct biomedical mappings have been manual, meaning the creation of new complex multi-ontology mappings has been reliant on efforts by domain experts. The ability to achieve complete constructs automatically could significantly decrease this effort, since correcting any missteps will require less time than creating the mapping from scratch.

Complex Multi-Ontology Matching presents as a unique problem essential to the full integration of ontologies in holistic scenarios of complex domains. CMOMgen is the first end-to-end strategy for CMOM that produces complete and semantically sound multi-ontology complex mappings, enabling the construction of alignments that link disjoint but related domains.

\section*{Acknowledgements}
This work was supported by FCT  - Fundação para a Ciência e Tecnologia, I.P. through project https://doi.org/10.54499/2022.11895.BD, the LASIGE Research Unit ref. UID/408/2025, and the INESC association ref. UIDB/50021/2020. It was also partially supported by the KATY project which has received funding from the European Union’s Horizon 2020 research and innovation program under grant agreement No 101017453, by the CancerScan project funded by the EU's HORIZON Europe research and innovation programme under grant agreement No 101186829 and by project 41, HfPT: Health from Portugal, funded by the Portuguese Plano de Recuperação e Resiliência.

\bibliographystyle{apalike} 
\bibliography{refs.bib}

\end{document}